\documentclass[
]{ceurart}

\usepackage{float}
\usepackage{listings}
\usepackage{booktabs}
\usepackage{subcaption}
\usepackage{graphicx}
\usepackage{caption}
\usepackage{cleveref}
\usepackage{listings}
\begin{document}

\copyrightyear{2024}
\copyrightclause{Copyright © 2024 for this paper by its authors. Use permitted under Creative Commons License Attribution 4.0 International (CC BY 4.0).}

\conference{EXPLIMED - First Workshop on Explainable Artificial Intelligence for the medical domain - 19-20 October 2024, Santiago de Compostela, Spain}

\title{Evaluating Machine Learning Models against \\
Clinical Protocols for Enhanced Interpretability \\
and Continuity of Care}

\author[1]{Christel Sirocchi}[%
orcid=0000-0002-5011-3068,
email=c.sirocchi2@campus.uniurb.it,
]
\cormark[1]
\fnmark[1]
\address[1]{Department of Pure and Applied Sciences, University of Urbino, Piazza della Repubblica 13, 61029, Urbino, Italy}

\author[1]{Muhammad Suffian}
\fnmark[1]

\author[1]{Federico Sabbatini}
\fnmark[1]

\author[1]{Alessandro Bogliolo}
\author[1]{Sara Montagna}

\cortext[1]{Corresponding author.}
\fntext[1]{These authors contributed equally.}

\definecolor{christelcolor}{rgb}{1.0, 0.0, 0.0}
\newcommand{\christel}[1]{\textcolor{christelcolor}{[#1]}}
\newcommand{\sara}[1]{\textcolor{teal}{[#1]}}
\newcommand{\Muhammad}[1]{\textcolor{blue}{[#1]}}
\newcommand{\csout}[1]{\christel{\sout{#1}}}

\begin{abstract}
In clinical practice, decision-making relies heavily on established protocols, often formalised as rules.  Concurrently, machine learning (ML) models, trained on clinical data, aspire to integrate into medical decision-making processes. However, despite the growing number of ML applications, their adoption into clinical practice remains limited. Two critical concerns arise, relevant to the notions of consistency and continuity of care: 
\emph{(a)} accuracy -- the ML model, albeit more accurate, might introduce errors that would not have occurred by applying the protocol; 
\emph{(b)} interpretability -- ML models operating as black boxes might make predictions based on relationships that contradict established clinical knowledge. 
In this context, the literature suggests using integrated ML models to reduce errors introduced by purely data-driven approaches and improve interpretability.
However, there is a lack of appropriate metrics for comparing ML models with clinical rules in addressing these challenges.

Accordingly, in this article, we first propose a metric to assess the accuracy of ML models with respect to the established protocol. Secondly, we propose an approach to measure the distance of explanations provided by two rule sets, with the goal of comparing the explanation similarity between clinical rule-based systems and rules extracted from ML models. 
The approach is validated by employing the Pima Indians Diabetes dataset, for which a well-grounded clinical protocol is available, by training two neural networks---one exclusively on data, and the other integrating knowledge.
Our findings demonstrate that the integrated ML model achieves comparable performance to that of a fully data-driven model while exhibiting superior relative accuracy with respect to the clinical protocol, ensuring enhanced continuity of care. Furthermore, we show that our integrated model provides explanations for predictions that align more closely with the clinical protocol compared to the data-driven model. 

\end{abstract}


\maketitle

\section{Introduction}
Machine learning (ML) has revolutionised various industries, from manufacturing to finance, and is now making its way into healthcare, a sector traditionally resistant to technological disruptions. ML has achieved remarkable performance in various domains of clinical medicine, outperforming human physicians in some cases and enabling the development of computer-aided diagnosis systems~\cite{piccialli2021survey}. 
With thousands of studies applying ML algorithms to medical data, only a handful have significantly contributed to clinical care, a stark contrast to the substantial impact ML has had in other industries. Indeed, only a few of these systems have been FDA-approved for healthcare use~\cite{benjamens2020state}.

Resistance to embrace ML in clinical settings can be attributed to the prevailing reliance on evidence-based clinical pathways, guidelines, and protocols as the foundation for clinical decision-making~\cite{clinton1994enhancing}. Adherence to established guidelines and practices is at the core of the consistency and continuity of care, defined as the degree to which a series of discrete healthcare events is experienced by people as coherent and consistent over time and across different healthcare providers~\cite{haggerty2003continuity}. Introducing novel decision-support systems offering alternative predictions and explanations may introduce variability among practices and practitioners, potentially compromising the quality and efficiency of care.

Novel ML models reporting superior performance compared to the current protocol might be found unsuitable for clinical use if they \emph{(a)} fail to correctly predict cases effectively managed by the protocol in place due to potential liabilities, or \emph{(b)} make predictions based on confounding variables and erroneous relationships that contradict established clinical knowledge~\cite{qian2021integrating}.
Therefore, in order to foster continuity of care when developing novel decision-support systems for healthcare, it is imperative not only to attain high overall accuracy but also to provide predictions and explanations adhering to current clinical guidelines. 
To this end, approaches integrating domain knowledge from clinical protocols into ML models have been proposed and proved effective~\cite{montagna2024hybrid}. 
However, metrics for evaluating the similarity of a novel model with respect to established protocols in terms of predictions and explanations are still lacking.

The main contribution of the manuscript is thus to introduce metrics capturing the adherence of a model to established protocols in terms of accuracy and explanation of its predictions. 
Specifically, we introduce the notions of \emph{relative accuracy} to quantify the proportion of samples correctly predicted by the model compared to those handled correctly by the existing protocol, and of \emph{explanation similarity} to quantify the degree of overlap between local explanations provided by the protocol and the ML model for the dataset instances. 

Through a comparison between a neural network model, trained solely on data, and a model incorporating domain knowledge encoded in a clinical protocol, we illustrate the potential of these metrics using the PIMA dataset~\cite{smith1988using}. While conventional performance metrics cannot definitively identify a superior model between the two, our proposed metrics reveal that the integrated model introduces fewer errors into the decision-making process and provides explanations that more closely mirror established practices. Consequently, these newly introduced metrics serve as valuable tools for identifying the ML model that better aligns with the protocol in place and is thus more suitable for integration into clinical practice in the prospect of continuity and consistency of care. 

Incorporating domain knowledge from clinical protocols into ML and developing metrics to evaluate the accuracy and interpretability of such models with respect to the protocol in place represent a pivotal step towards overcoming the limitations of ML and facilitating its seamless integration into medical practice.

\clearpage
\section{Background and previous work}


As medical decision-making becomes increasingly complex due to the development of new therapies and diagnostics, as well as the accumulation of health records, ML has emerged as a promising tool to support medical decision-making processes for its ability to model complex interactions between features~\cite{obermeyer2017lost}. However, the application of ML in healthcare presents several challenges, primarily related to the quantity, quality, and composition of clinical data, as well as a lack of explainability and limited robustness~\cite{leiser2023medical}.
To address these limitations, the literature reports on various integrative approaches that leverage multiple models, data sources, and prior knowledge. 
A notable advancement is the paradigm of \textit{Informed Machine Learning}, which integrates data and prior knowledge derived from independent sources to strike a balance between model complexity and effectiveness~\cite{von2021informed,sirocchi2024medical}. This approach 
has gained attention in the medical domain, where structured knowledge is abundant but data is often limited and noisy.
Recent contributions in this area have provided taxonomies of integration strategies applied to the healthcare sector, with a focus on the integration of ML with rule-based expert systems, highlighting that integration can be beneficial across all phases of the ML pipeline, from data preprocessing and feature engineering, to model learning and output evaluation~\cite{leiser2023medical,kierner2023taxonomy,sirocchi2024medical}. 
Particular emphasis is placed on strategies incorporating prior knowledge into the model's loss function, often through regularisation or penalty terms quantifying inconsistencies or violations concerning the knowledge base. This approach has shown promising results in clinical applications, enhancing model performance, robustness, and interpretability~\cite{leiser2023medical}. 

Despite these advancements, there remains a critical need of metrics to evaluate the resulting hybrid models against the knowledge base to measure adherence, and against the data-driven counterpart to quantify knowledge injection. For a comprehensive comparative analysis, such metrics must evaluate both accuracy and interpretability.
This study aims to address this gap by proposing metrics assessing model adherence to a knowledge base in terms of performance and explainability, with immediate applications in evaluating hybrid models in clinical settings. 

\subsection{Evaluating model performance}

A plethora of scores have been proposed to gauge the correctness of the predictions with respect to the ground truth. In classification tasks, accuracy emerges as an intuitive metric, quantifying the proportion of correctly classified instances. Sensitivity (recall) and specificity measure the proportion of correctly identified true positives and true negatives, respectively, while precision evaluates the ratio of true positives over positive predictions. F1-score, the harmonic mean of precision and recall, balances both metrics. The area under the receiver operating characteristic curves provides a comprehensive view of model performance across different thresholds but is less interpretable to some stakeholders. Overall, accuracy and F1-score are the most popular metrics but may yield overly optimistic results with imbalanced data. Recently, Matthew's correlation coefficient has gained prominence in biomedical data analysis 
for yielding more trustworthy results in imbalanced datasets~\cite{chicco2020advantages}.

Determining the most suitable statistical metric remains challenging, with no consensus reached~\cite{chicco2020advantages}. Comparative analyses often leverage a diverse array of metrics for a comprehensive evaluation, with the choice of the most appropriate metric contingent upon the specific case at hand. In clinical contexts, for instance, recall often takes precedence, as the cost (risk) associated with false negatives outweighs that of false positives (as a positive result typically leads to additional, more precise tests, unlike a negative one). However, in such contexts where long-standing guidelines are in place, it is also crucial to evaluate ML models with respect to the protocol as well as the ground truth, as mistakes introduced by the model also carry high costs. 

\subsection{Evaluating model explanations}\label{subsec:xai}
As ML architectures become increasingly complex, there arises a pressing need to bridge the gap between the opaque nature of these models and human comprehension, especially in domains like healthcare where transparency and interpretability are essential~\cite{sokol2020explainability}. Addressing this challenge, the field of eXplainable Artificial Intelligence (XAI) has emerged, developing tools aimed at providing human-understandable explanations for AI-driven decisions, thereby fostering transparency, trust, and collaboration between human expertise and computational intelligence~\cite{yang2022explainable}.
%
%
%
XAI employs various techniques to provide reasoning on ML decisions, mainly operating on two levels: local and global. In the former, individual model predictions are analysed, while in the latter the overall behaviour of the model is analysed to identify patterns and relationships in the data. 
%
Among XAI techniques, feature importance methods have emerged as influential for identifying important variables. 
Additionally, example-based explanations offer insights by presenting similar instances in the dataset that influenced the predictions.

Rule extraction techniques translate the ML models into human-understandable rules or decision trees which provide insights into the overall behaviour of the model across the entire dataset~\cite{calegari2020integration,SKESKISLR2024}. Moreover, a rule applicable to a given data instance indicates the conditions that were satisfied to produce the corresponding outcome, offering explanations at the local level.
%
Several rule extraction algorithms exist in the literature. 
The Rule Extraction From Neural Network Ensemble (REFNE)~\cite{REFNE} was initially developed for extracting symbolic rules from neural network ensembles. However, its accuracy decreases when the data is highly complex or nonlinear.
%
C4.5Rule-PANE~\cite{vilone2021quantitative} utilises the C4.5 rule induction algorithm to extract if-then rules from neural networks and, like other tree-based algorithms, is susceptible to over-fitting.
TREPAN~\cite{craven1996extracting} constructs a decision tree by querying the underlying network to determine output classes. However, it often extracts suboptimal rule sets and requires binary inputs.

%
%
Decision trees, particularly Classification and Regression Trees (CART)~\cite{breiman2017classification}, remain one of the most prominent approaches in rule extraction.  CART constructs a binary tree structure which is then translated into human-readable rules by converting each possible path from the root to the leaves into an if-then rule. 
Its strengths are simplicity, interpretability, and ability to handle both categorical and numerical data effectively.
%
Several other rule-extraction algorithms exist, as well as software libraries dedicated to knowledge extraction, e.g., the PSyKE platform~\cite{sabbatini2021design}, providing a unified software framework supporting various rule-extraction methods \cite{craven1996extracting,breiman2017classification,craven1994using,huysmans2006iter,gridex-extraamas2021}.
Several evaluation metrics are documented in the literature to assess the quality of extracted rule sets. Among these metrics, the number of rules and average rule length reflect attributes of the explainability of the rule extractor. The other metrics -- completeness, correctness, fidelity, robustness, and coverage -- serve as general validation factors applicable to any rule extractor method. These metrics primarily analyse properties of the rules as global explanations for the model, offering a coarse-grained evaluation. Less attention is given to metrics assessing rules as local explanations for dataset instances, which would offer a more nuanced and context-aware evaluation, particularly relevant in the clinical setting.
Furthermore, while most metrics evaluate the properties of a single rule set, there is a noticeable scarcity of similarity measures comparing multiple rule sets. Existing literature reports similarity metrics over dataset instances (e.g., Jaccard~\cite{jaccard}, Cosine~\cite{cosine}, Dice~\cite{dice}) or similarities between rule-based knowledge bases (e.g., XNOR). However, there is a lack of similarity metrics over rule-based local explanations aggregated across data instances to provide global measures of similarity between rule sets.

\section{Methods}

This section is structured as follows: Section~\ref{sec:data} details the dataset and domain knowledge used in our case study, Section~\ref{sec:IML} describes the machine learning model that integrates this domain knowledge, while Section~\ref{sec:metrics} introduces the metrics for accuracy and explainability used to evaluate the integrated model against clinical knowledge and a data-driven model.

\subsection{Dataset and domain knowledge}\label{sec:data}
In this work, we present our investigations involving the Pima Indians Diabetes dataset, originally compiled by the National Institute of Diabetes and Digestive and Kidney Diseases from a study of the Pima Indian population, known for its notably high incidence of diabetes~\cite{smith1988using}. The dataset comprises 768 medical profiles of women aged 21 and above, who underwent an Oral Glucose Tolerance Test (OGTT) to measure their glucose and insulin levels at two hours. The target variable is binary, indicating a diabetes diagnosis within five years. Table~\ref{tab:pima} reports the 8 input features available in the dataset. Missing values are present in the attributes $I_{120}$ (48.70\%), $ST$ (29.56\%), $BP$ (4.55\%), $BMI$ (1.43\%), and $G_{120}$ (0.65\%), and were imputed in this work with the median value of the respective variable, as reported in the literature~\cite{kibria2022ensemble}. Further details about this dataset can be found in Table~\ref{tab:pima}. 

\begin{table}[h]
\caption{Pima Indians Diabetes dataset}\label{tab:pima}%
\begin{tabular}{@{}lcl@{}}
\toprule
Feature name & Code & Description\\ 
\midrule
Pregnancies & & Number of times pregnant\\
Glucose & $G_{120}$ & 2-hour plasma glucose concentration in OOGT in $mg/dL$
\\
Blood Pressure & $BP$ & Diastolic blood pressure in $mm Hg$\\
Skin Thickness & $ST$ & Triceps skin-fold thickness in $mm$\\
Insulin & $I_{120}$ & 2-hour serum insulin in $\mu U/mL$\\
Body mass index & $BMI$ & Body mass index as weight/(height)$^2$ in $kg/m^2$\\
Diabetes Pedigree Function & $DPF$ & Likelihood function of diabetes based on family history~\cite{smith1988using}\\
Age & & Age in years\\
\midrule
\end{tabular}
\end{table}

\noindent Public health guidelines on type-2 diabetes risks report that individuals with a high $BMI$ ($\geq$ 30) and high blood glucose level ($\geq$ 126) are at severe risk for diabetes, while those with normal $BMI$ ($\leq$ 25) and low blood glucose level ($\leq$ 100) are less likely to develop diabetes. These guidelines have been utilised to design rules~\cite{kunapuli2010online} expressed as logic predicates (see \Cref{tab:rules}).

\begin{table}[t]
\caption{Knowledge base for predicting risk of type-2 diabetes as formalised by Kunapuli et al. 2020.}
\label{tab:rules}
\renewcommand{\arraystretch}{1.5}
\begin{tabular}{|lll|}
\hline
Rule 1 & & $(BMI \geq 30) \land (G_{120} \geq 126) \implies \text{diabetes}$ \\
Rule 2 & & $(BMI \leq 25) \land (G_{120} \leq 100) \implies \text{healthy}$ \\
\hline
\end{tabular}
\end{table}

\subsection{Integrated ML model}\label{sec:IML}
The hybrid ML model examined in this study, herein denoted as KB-ML, integrates domain knowledge in the loss function. Specifically, KB-ML is a neural network for binary classification trained using a custom loss function that assigns greater weight to samples accurately predicted by the clinical guidelines represented by the two logic predicates in Table~\ref{tab:rules}.
Formally, let $\mathcal{D}$ denote a dataset comprising $n$ instances each represented by $\boldsymbol{x}_i$, where $i$ ranges from 1 to $n$. 
Three $n \times 1$ vectors $\boldsymbol{y}$, $\boldsymbol{p}$, and $\boldsymbol{r}$ can be defined. Vector $\boldsymbol{y}$ contains the ground-truth binary labels, with each element denoted as $y_i$ and representing the expected outcome for instance $\boldsymbol{x}_i$. Vector $\boldsymbol{p}$ contains the probability of the outcome belonging to the positive class predicted by the neural network, with each element $p_i$ corresponding to $\boldsymbol{x}_i$. Finally, vector $\boldsymbol{r}$ contains the predictions according to the rules in Table~\ref{tab:rules}, i.e., each element $r_i$ takes value 1 if $\boldsymbol{x}_i$ satisfies the conditions of the first rule, 0 if it satisfies the second rule, and N/A otherwise.
Then, the Custom Total Loss (CTL) for the integrated model is computed as:
\begin{equation}
\text{CTL}(\boldsymbol{y}, \boldsymbol{p}, \boldsymbol{r}, \alpha) = \frac{1}{n} \sum_{i=1}^{n} \text{CSL}(y_i, p_i, r_i, \alpha),
\label{eq:CTL}
\end{equation}
where $\alpha$ is the scaling factor controlling the influence of the additional loss term, CSL is the custom binary cross-entropy loss for a single sample defined as
\begin{equation}
\text{CSL}(y_i, p_i, r_i, \alpha) =
\begin{cases}
L(y_i, p_i) & \text{if } r_i \neq y_i \\
(\alpha + 1) L(y_i, p_i) & \text{if } r_i = y_i
\end{cases}
\end{equation}
and $L$ is the standard binary cross-entropy loss for a single sample
\begin{equation}
L(y_i, p_i) = - \left[ y_i \cdot \log(p_i) + (1 - y_i) \cdot \log(1 - p_i) \right].
\end{equation}
    
\subsection{Proposed evaluation metrics}\label{sec:metrics}

\subsubsection{Relative accuracy}
Performance metrics can be redefined to evaluate adherence to accurate predictions set by the rules, quantifying errors introduced by the model in comparison to the reference protocol.
As in Section~\ref{sec:IML}, consider $\mathcal{D}$ as a dataset consisting of $n$ samples represented by $\boldsymbol{x}_i$, where $i$ ranges from 1 to $n$, and let $r_i$ denote the prediction made by a clinical protocol for each $\boldsymbol{x}_i$. Additionally, let $\hat{y}_i$ represent the binary prediction provided by a ML model for $\boldsymbol{x}_i$.
Relative Accuracy (RA) can be defined as the fraction of samples correctly predicted by the protocol that are also correctly predicted by the model:
\begin{equation}
\text{RA} = \frac{\left| \{\boldsymbol{x}_i : \boldsymbol{x}_i \in \mathcal{D} \wedge r_i = y_i = \hat{y}_i\} \right|}{\left| \{\boldsymbol{x}_i : \boldsymbol{x}_i \in \mathcal{D} \wedge r_i = y_i \} \right|},
\end{equation}
where $ \left| \cdot \right| $ denotes the cardinality of a set. 
Similarly, the relative counterparts for other performance metrics, such as Relative sensitivity or Recall ($ \text{RR} $) and Relative Specificity ($ \text{RS} $) with respect to a given class $ c $, can be defined as follows:
\begin{equation}
\text{RR} = \frac{\left| \{\boldsymbol{x}_i : \boldsymbol{x}_i \in \mathcal{D} \wedge r_i = y_i = \hat{y}_i = c\} \right|}{\left| \{\boldsymbol{x}_i : \boldsymbol{x}_i \in \mathcal{D} \wedge r_i = y_i = c\} \right|},
\end{equation}
\begin{equation}
\text{RS} = \frac{\left| \{\boldsymbol{x}_i : \boldsymbol{x}_i \in \mathcal{D} \wedge y_i \neq c \wedge \ r_i \neq c \wedge \ \hat{y}_i \neq c\} \right|}{\left| \{\boldsymbol{x}_i : \boldsymbol{x}_i \in \mathcal{D} \wedge y_i \neq c \wedge \ r_i \neq c\} \right|}.
\end{equation}
This evaluation does not account for samples where the protocol makes errors or fails to provide a prediction, requiring additional performance metrics for a comprehensive assessment. 


\subsubsection{Explanation similarity}\label{subsub:expSimilarity}

Applying XAI in clinical settings requires proper evaluation to ensure the explanations are both technically sound and clinically useful. Rule sets extracted from ML models provide valuable insights into model behaviour. Notably, rules extracted from different ML models can emphasise different variables, even when predicting similar outcomes. Therefore, it is crucial to assess the similarity of explanations provided by rules approximating predictors to those offered by a specified reference protocol. This evaluation helps determine which explanation aligns more closely with the clinical protocol in use and better reflects clinical expertise.  

    

\begin{figure}[!ht]
\centering 
\includegraphics[width=\textwidth]{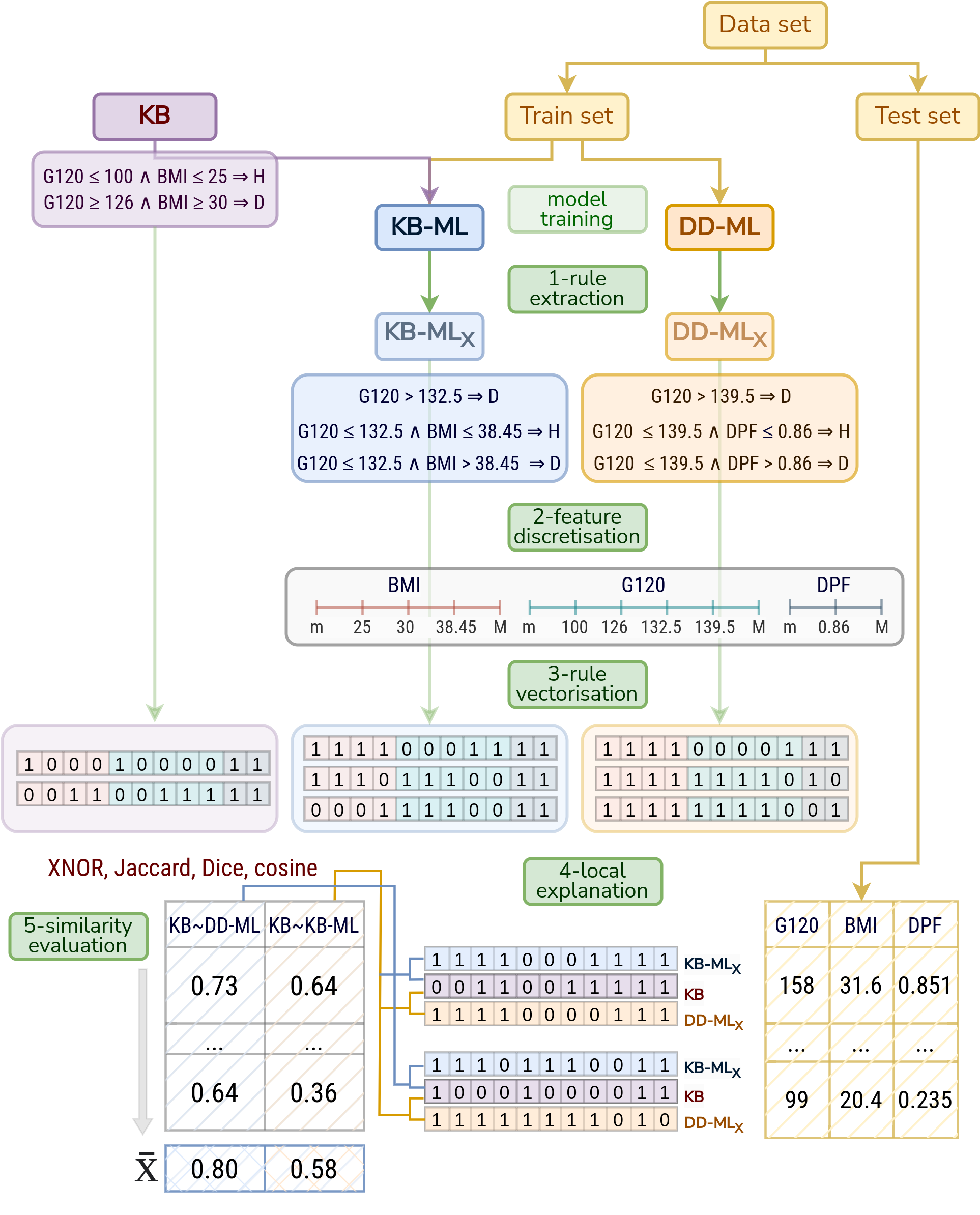}
\caption{Diagram illustrating the proposed approach for assessing explanation similarity between a knowledge base (KB) and the rule sets KB-ML$_X$ and DD-ML$_X$, derived from rule extraction from a data-driven model (DD-ML) and an integrated model (KB-ML), predicting diabetes (D) or healthy (H) outcomes for instances of the Pima Indians Diabetes dataset.}
\label{fig:diagram}
\end{figure}

A novel explanation similarity strategy is here proposed to estimate the similarity of explanations from rule-based predictors, whether extracted from black-box models or built on clinical knowledge. This method allows for comparing explanations from integrated and data-driven models with those provided by a clinical protocol, to verify which aligns better. A diagram summarising the approach is shown in Figure~\ref{fig:diagram}. The method entails the following steps.

\begin{description}
\item[1. Rule extraction] symbolic knowledge is extracted from black-box predictors trained on a given dataset and represented as rule sets that are both human- and machine-interpretable and can provide explanations to predictions in the form of first-order logic clauses.
\item[2. Feature discretisation] the features of the dataset are discretised according to the thresholds found in the rules of the considered rule sets. This involves collecting all thresholds associated with each feature and discretising the feature into intervals, accordingly. 
\item[3. Rule vectorisation] each rule is assigned a vector representing the feature space, where each element corresponds to an interval of a feature and is assigned a value of 1 if the corresponding feature and interval satisfy the rule, and 0 otherwise.
\item[4. Local explanation] for every rule set, and for each sample in the data set, the rule satisfied by the sample is identified and the corresponding vector is assigned to the sample. 
\item[5. Similarity calculation] the similarity between two rule sets is obtained by computing, for each sample, the similarity between the vectors obtained from the two rule sets, and averaging across all samples, while the similarity among \textit{more than two rule sets} is obtained by calculating the similarity between each pair of rule sets, and averaging all scores.
\end{description}


\noindent Formally, let $\mathcal{D}$ represent a dataset comprising $n$ samples denoted by $\boldsymbol{x}_s$, where $s$ ranges from 1 to $n$. Each sample is described by $m$ input features, labelled as $v_1, v_2, \ldots, v_m$. Here, $x_s^k$ represents the value of feature $v_k$ in the instance $\boldsymbol{x}_s$. For each input $\boldsymbol{x}_s$, $y_s$ denotes the corresponding outcome. $D_x$ and $D_y$ denote the domains of the inputs and outputs, respectively:

$$
\Bigl( \boldsymbol{x}_s \in D_x \Bigr) \wedge \Bigl( y_s \in D_y \Bigr), \quad \forall s = 1, 2, \dots, n.
$$

\paragraph{Rule extraction.}

Let us consider a predictive function $\mathcal{F}$ 
$$\mathcal{F} : D_x \rightarrow D_y, \quad \mathcal{F}(\boldsymbol{x}_s) = \hat{y}_s,$$
where $\hat{y}_s$ is the value predicted by $\mathcal{F}$ for the instance $\boldsymbol{x}_s$. Then, a rule set $\mathcal{R}$ mapping instances to outputs and approximating the input-output relationship of $\mathcal{F}$ can be obtained by analysing $\mathcal{F}$.
%
Let $\mathcal{P}\;$ be a set of $p$ rule sets, either obtained from predictive functions by rule extraction or available from domain knowledge, which we aim to compare:
$$
\mathcal{P} = \{ \mathcal{R}^1, \mathcal{R}^2, \ldots, \mathcal{R}^p \},
$$
where 
$$
\mathcal{R}^i : D_x^i \subseteq D_x \rightarrow D_y \quad \forall \; i = 1, 2, \dots, p.
$$ 
Each rule set consists of rules denoted by $R$. For instance, if rule set $i$ comprises $q$ rules, then $\mathcal{R}^i = \{ R_1^i, R_2^i, \ldots, R_q^i \}$. 
Each rule $R_j^i$ in rule set $\mathcal{R}^i$ is represented as a tuple $(C_j^i, \hat{Y}_j^i)$, where $C_j^i$ constitutes a set of $t$ conditions $\{ c_{j1}^i, c_{j2}^i, \ldots, c_{jt}^i \}$ and $\hat{Y}_j^i$ represents the outcome associated to that rule. Each condition $c_{jh}^i$ can be expressed by a tuple $(v_{jh}^i, l_{jh}^i, u_{jh}^i)$, where $v_{jh}^i$ is the variable included in the condition, and $l_{jh}^i$ and $u_{jh}^i$ are the lower and upper bounds for the condition. If a condition is defined over a discontinuous interval, it is separated into distinct conditions. If a condition is of the type \textit{less than} or \textit{greater than}, the lower or upper bound is replaced with the minimum or maximum value of the variable for that feature in the dataset.
\newline 

For instance, in the considered case study where $BMI$ is in the range [18, 67] and $G_{120}$ in [67, 199], the rule set $\mathcal{R}^1$ presented in Table~\ref{tab:rules} is defined as $\mathcal{R}^1 = \{ R_1^1, R_2^1 \}$, where $R_1^1 = \{(C_1^1, \text{diabetes})\}$ with $C_1^1 = \{(BMI, 30, 67), (G_{120}, 126, 199)\}$ and $R_2^1 = \{(C_2^1, \text{healthy})\}$ with $C_2^1 = \{(BMI, 18, 25), (G_{120}, 44, 100)\}$.

\paragraph{Feature discretisation.}

For the set of predictors $\mathcal{P}$, we define the set of thresholds $\mathcal{T}$ as:
$$
\mathcal{T} = \{ T(v_1), T(v_2), \ldots, T(v_m)\},
$$
where $T(v_k)$ denotes the set of thresholds (upper and lower bounds) found in the conditions of the rules in $\mathcal{P}$ on feature $v_k$:
$$
T(v_k) =  \bigcup_{\mathcal{R}^i \in \mathcal{P}} \left( \bigcup_{R_j^i \in \mathcal{R}^i} \left( \bigcup_{(C_j^i,\, \hat{Y}_j^i) \, \in \, R_j^i} \left( \bigcup_{c_{jh}^i \in C_j^i} \{ l_{jh}^i,\, u_{jh}^i \, |\, v_{jh}^i = v_k \} \right) \right) \right).
$$
 
If feature $v_k$ never occurs in any conditions of $\mathcal{P}$, then $|T(v_k)| = 0$. Each set $T(v_k)$ can be represented as an ordered set of thresholds retrieved from rule conditions as detailed above:
$$
T(v_k) = ( \phi_{k1}, \phi_{k2}, \ldots, \phi_{kz} ) , \quad \phi_{k1} < \phi_{k2} < \ldots < \phi_{kz}.
$$

\paragraph{Rule vectorisation.}

For each rule $R_j^i$, we define a set of binary vectors $\mathcal{I}_j^i$:
$$
\mathcal{I}_j^i = \{ I_j^i(v_1), \; I_j^i(v_2), \; \ldots, \; I_j^i(v_m)\},
$$
where $I_j^i(v_k)$ is a binary vector representing intervals for variable $v_k$. If $v_k$ is not present in any rule, i.e., $|T(v_k)| = 0$, then this vector has zero length. Otherwise, the vector has length $|T(v_k)|-1$, and the $r$-th element of the vector corresponds to the interval $[\phi_{kr}, \phi_{k(r+1)}]$.
The $r$-th element of the vector is set to $1$ if the values in the corresponding interval meet all conditions on that variable for the considered rule, or if no conditions on that variable are specified in the rule. Otherwise, the element is set to $0$:
\begin{equation*}
I_j^i(v_k)[r] = 
\begin{cases}
1 & \text{if } [\phi_{kr}, \phi_{k(r+1)}] \subseteq [l_{jh}^i, u_{jh}^i] ~~ \forall (v_{jh}^i, l_{jh}^i, u_{jh}^i) \in R_j^i : v_{jh}^i = v_k, \\
1 & \text{if }
v_{jh}^i \neq v_k ~~ \forall (v_{jh}^i, l_{jh}^i, u_{jh}^i) \in R_j^i, \\
0 & \text{otherwise}.
\end{cases}
\label{eq:vector}
\end{equation*}
Then vector $V_j^i$ is obtained from $\mathcal{I}_j^i$ by concatenating all vectors into a single one: 
$$
V_j^i = I_j^i(v_1)I_j^i(v_2) \ldots I_j^i(v_m).
$$

\paragraph{Local explanation.}

Let $\mathcal{D}_R$ be the subset of instances in $\mathcal{D}$ for which each of the considered rule sets can provide a prediction, i.e.,
$$
\mathcal{D}_R = \left\{ \boldsymbol{x}_s \; \middle| \; \boldsymbol{x}_s \in D_x \land \, \boldsymbol{x}_s \in \bigcap_{i=1}^{p} D_x^i \right\}.
$$ 
Then, $\rho_s^i$ is the set of rules in $\mathcal{R}^i$ such that the instance $\boldsymbol{x}_s$ satisfies all the conditions of the rule:

$$
\rho_s^i = \bigcup_{R_j^i \in \mathcal{R}^i} \left\{ R_j^i \; \middle| \; v_{jh}^i = v_k \land x_s^k \in [l_{jh}^i, u_{jh}^i], \forall \; c_{jh}^i \in R_j^i \right\}.
$$ 

Here we assume, for each rule set in $\mathcal{P}$, that each instance of the dataset satisfies all conditions for only one rule, i.e. $|\rho_s^i|  = 1 \; \forall \; \boldsymbol{x}_s \in \mathcal{D}_R$.
%
The vector corresponding to the rule in $\rho_s^i$ is assigned to $\boldsymbol{x}_s$ and denoted as $V^i(\boldsymbol{x}_s)$. This provides a vectorised representation of the explanation offered by rule set $\mathcal{R}^i$ for the data instance $\boldsymbol{x}_s$.

Without loss of generality, the rule vectorisation and local explanation procedure can also be applied to categorical variables. Instead of intervals defined by thresholds, we have vectors representing subsets of possible categorical values, and conditions are verified by set inclusion.

\paragraph{Similarity evaluation.}
Let $S(V^1,V^2)$ be a similarity function on two binary vectors $V^1$ and $V^2$. The similarity $\mathcal{S}(\mathcal{R}^1,\mathcal{R}^2)$ for two rule sets $\mathcal{R}^1$ and $\mathcal{R}^2$ in $\mathcal{P}$ can then be computed as 
\begin{equation}
\mathcal{S}(\mathcal{R}^1,\mathcal{R}^2,S,\mathcal{D}_R) = \frac{1}{|\mathcal{D}_R|} \sum_{\boldsymbol{x}_s \in \mathcal{D}_R} S(V^1(\boldsymbol{x}_s), V^2(\boldsymbol{x}_s)).
\end{equation}

The similarity among more than two sets is computed by calculating the pairwise similarity between each pair of rule sets and then averaging across all rule sets. For a set $\mathcal{P}$ of $p$ rule sets the similarity is computed as:

\begin{equation}
\mathcal{S}(\mathcal{P},S,\mathcal{D}_R) = \frac{2}{p(p-1)} \frac{1}{|\mathcal{D}_R|} \sum_{f=1}^{p} \sum_{g=1}^{f-1} \sum_{\boldsymbol{x}_s \in \mathcal{D}_R} S(V^f(\boldsymbol{x}_s), V^g(\boldsymbol{x}_s)).
\end{equation}

To compute the similarity of two binary vectors $V^1$ and $V^2$ of length $w$, various similarity metrics $S$ are available in the literature.

\subparagraph{XNOR similarity} considers matching and non-matching elements:
\begin{equation}\label{eq:xnor}
    \text{XNOR}(V^1, V^2) = \frac{\sum_{i=1}^{w} \delta(V^1[i], V^2[i])}{w}
\end{equation}
where $\delta(V^1[i], V^2[i])$ equals $1$ if $V^1[i]=V^2[i]$ and $0$ otherwise.

\subparagraph{JACCARD similarity} considers the intersection over the union of elements in both vectors:
\begin{equation}\label{eq:jaccard}
    \text{JACCARD}(V^1, V^2) = \frac{\sum_{i=1}^{w} V^1[i] \cdot V^2[i]}{\sum_{i=1}^{w} V^1[i] + \sum_{i=1}^{w} V^2[i] - \sum_{i=1}^{w} V^1[i] \cdot V^2[i]}
\end{equation}
where $\sum_{i=1}^{w} V^1[i] \cdot V^2[i]$ counts the elements that are $1$ in both vectors (intersection), while $\sum_{i=1}^{w} V^1[i] + \sum_{i=1}^{w} V^2[i]$ counts the elements that are $1$ in either rule vector (union).

\subparagraph{COSINE similarity} computes the cosine of the angle between the vectors:
\begin{equation}\label{eq:cosine}
    \text{COSINE}(V^1, V^2) = \frac{\sum_{i=1}^{w} V^1[i] \cdot V^2[i]}{\sqrt{\sum_{i=1}^{w} V^1[i]^2} \cdot \sqrt{\sum_{i=1}^{w} V^2[i]^2}}
\end{equation}
where $\sqrt{\sum_{i=1}^{w} V^1[i]^2} \cdot \sqrt{\sum_{i=1}^{w} V^2[i]^2}$ computes the product of the magnitudes of the vectors. 

\subparagraph{DICE similarity} divides twice the number of matching elements by the number of elements:
\begin{equation}\label{eq:dice}
    \text{DICE}(V^1, V^2) = \frac{2 \cdot \sum_{i=1}^{w} V^1[i] \cdot V^2[i]}{\sum_{i=1}^{w} V^1[i] + \sum_{i=1}^{w} V^2[i]}
\end{equation}

\subsection{Evaluation strategy}

The study conducted a comparison between two neural networks trained on the Pima Indians Diabetes dataset. One model, termed the data-driven model (DD-ML), was exclusively trained on data, while the other, referred to as the integrated or knowledge-based model (KB-ML), was trained with a custom loss function incorporating knowledge from a knowledge base (KB), as detailed in Section~\ref{sec:IML}.
%
Both neural networks were designed as feed-forward models, comprising three fully connected layers: two hidden layers with rectified linear unit activation functions and an output layer with a sigmoid activation function. DD-ML was trained using binary cross-entropy loss, whereas KB-ML employed a customised loss function defined in Eq.~\ref{eq:CTL} with parameter $\alpha$, tuning the contribution of KB to model learning, ranging from 0.5 to 4 at intervals of 0.5. All neural networks were trained with a batch size of 20 for 25 epochs. 

In all experiments, data was divided into training and testing sets using a 10-times 10-fold stratified cross-validation approach~\cite{bouckaert2004evaluating}. The performance and explainability metrics computed for the integrated model were evaluated against the corresponding metrics for the data-driven model using paired Student-t tests with the Nadeau and Bengio correction~\cite{nadeau1999inference}. 
%
Performance evaluation encompassed a range of metrics, including Accuracy (A), F1-score (F1), Recall (R), Precision (P), Balanced Accuracy (BA), the Area Under the Receiver Operating Characteristic Curve (ROC AUC), and Matthews Correlation Coefficient (MCC). Moreover, the Relative Accuracy (RA), Sensitivity (RR) and Specificity (RS) metrics herein introduced were computed for all models.

Interpretable models approximating the predictions of the neural networks were obtained by rule extraction using CART~\cite{breiman2017classification} available from the PSyKE library~\cite{sabbatini2022symbolic}.
%
Rule sets were extracted from DD-ML and KB-ML (trained with the tuning parameter $\alpha$ set to 1.5) and denoted as DD-ML$_X$ and KB-ML$_X$, respectively. Thus, each experiment yields three rule sets: KB, which formalises the clinical protocol; DD-ML$_X$, which approximates the data-driven model; and KB-ML$_X$, which approximates the integrated model. The maximum number of leaves, and thus rules, in the CART rule-extraction process, varied from 2 to 12. The fidelity of the obtained rule set was evaluated in terms of accuracy and F1-score with respect to the black-box model.
%

The proposed explanation similarity metrics (leveraging XNOR, Dice, Jaccard, and Cosine similarity) were computed between DD-ML$_X$ and KB, and between DD-ML$_X$ and KB on two subsets of the dataset. Initially, explanation similarity metrics were computed over samples for which all considered predictors (KB, DD-ML, KB-ML) could make predictions, thus excluding samples not handled by the protocol. Subsequently, explanation similarity metrics were computed over samples for which all considered models made correct predictions. 
Finally, the explanation similarity metrics were utilised to gauge the robustness of explanations. A comparison was made among the 100 instances of the KB-ML model trained over the 10-times 10-fold cross-validation. A 100x100 similarity matrix was generated, computing pairwise explanation similarity with XNOR operation between each pair of model instances. The similarities were then averaged across all elements of the matrix. The same process was repeated for DD-ML.

\section{Results and discussion}

\subsection{Relative accuracy evaluation}
\begin{figure}[!ht]
    \centering
    \begin{subfigure}{0.54\textwidth}
        \centering
        \includegraphics[width=\textwidth]{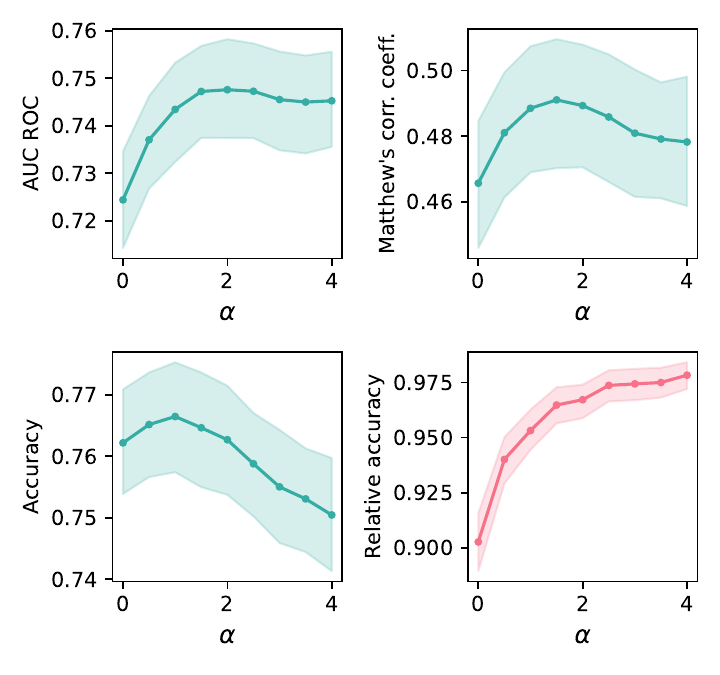}
        \caption{}
    \end{subfigure}\hfill
    \begin{subfigure}{0.31\textwidth}
        \centering
        \includegraphics[width=\textwidth]{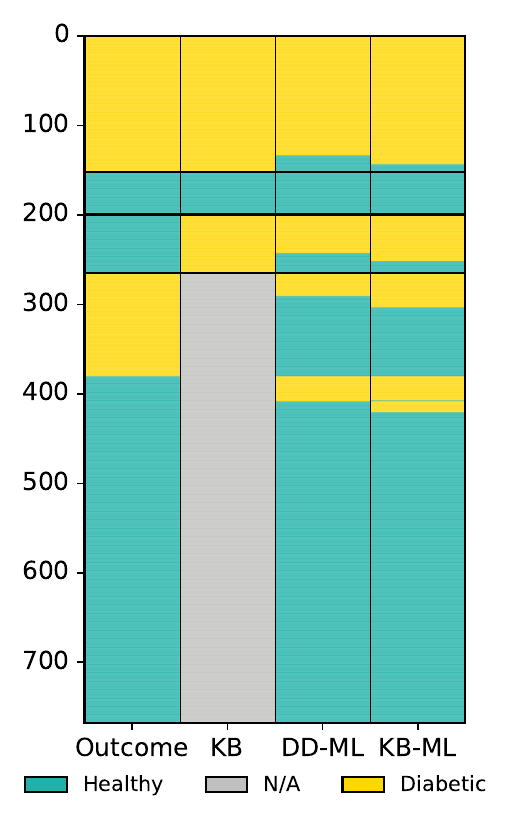}
        \caption{}
    \end{subfigure}
    \caption{(a) Performance metrics for the integrated model (KB-ML) with parameter $\alpha$, ranging from 0 to 4, averaged over 100 iterations with 95\% confidence intervals. For $\alpha = 0$, the model corresponds to the fully data-driven model (DD-ML). 
    (b) Comparison of true labels, outcomes of the clinical protocol (KB) and prediction of the two models averaged over the 10 folds of the cross-validation.}
    \label{fig:ra_alpha}
\end{figure}

This integration of domain knowledge, modulated by the parameter $\alpha$, influences the model's performance, which varies with respect to $\alpha$ as shown in Figure~\ref{fig:ra_alpha}a. For standard metrics, the performance increases, peaking between $\alpha$ values of 1 and 1.5, subsequently declining for A and MCC, while stabilising for ROC. This trend suggests that while the introduced learning bias given by the protocol could be beneficial, excessive bias might impede the learning process, leading to decreased accuracy that falls below that of the data-driven model for $\alpha$ greater than 2. The proposed RA metric increases with $\alpha$, effectively detecting the reduction of errors introduced by the integrated model with respect to the reference model.
For values of $\alpha$ around 1.5, optimal scores of standard metrics are achieved, as well as improved RA. This evaluation highlights the need of tuning integration to maximise adherence without compromising performance.  

A comprehensive array of metrics comparing the data-driven model with the integrated model at $\alpha$ equal to 1.5, along with relative p-values indicating statistical significance, are reported in Table~\ref{tab:res}. The integrated model yields superior scores across all metrics except precision, with statistical significance observed for BA, ROC, and R. Nonetheless, precision significantly decreases, and improvements in MCC, F1, and A lack statistical significance. Therefore, it remains challenging to conclusively state that one model is superior to the other. However, the RA metric significantly improved from 0.90 to 0.97, driven by the increased RR (since RS is maximal for both models). These findings highlight the greater alignment with the clinical protocol, also seen in Figure~\ref{fig:ra_alpha}b, making the integrated model preferable overall, and demonstrate the role of the proposed metrics in facilitating this assessment.

\begin{table}[!ht]
    \caption{Average of performance metrics for the data-driven model (DD-ML), and the integrated model (KB-ML), trained using 10-times 10-fold cross-validation. Performance differences are evaluated using the paired Student-t test with the Nadeau and Bengio correction, with the resulting p-values reported. Bold values highlight significant performance differences between the models at a 0.05 significance level.}\label{tab:res}
    \centering
    \begin{tabular}{r|ccccccccccc}
    \hline
    ~ & MCC & F1 & A & BA & ROC & P & R & RA & RR & RS & \\ \hline
    DD-ML & 0.466 & 0.729 & 0.762 & 0.724 & 0.724 & \textbf{0.684} & 0.599 & 0.903 & 0.873 & 1.000\\ 
    KB-ML & 0.491 & 0.743 & 0.765 & \textbf{0.747} & \textbf{0.747}  & 0.657 & \textbf{0.689} & \textbf{0.965} & \textbf{0.954} & 1.000\\ \hline
    p-value & 0.153 & 0.139 & 0.408 & 0.045 & 0.045 & 0.046 & 0.001 & 0.001 & 0.001 & - \\ \hline
    \end{tabular}
\end{table}

\subsection{Explanation similarity evaluation}
\begin{figure}[p!]
 \begin{center}
    \begin{subfigure}{\textwidth}
        \centering
        \includegraphics[width=.9\linewidth]{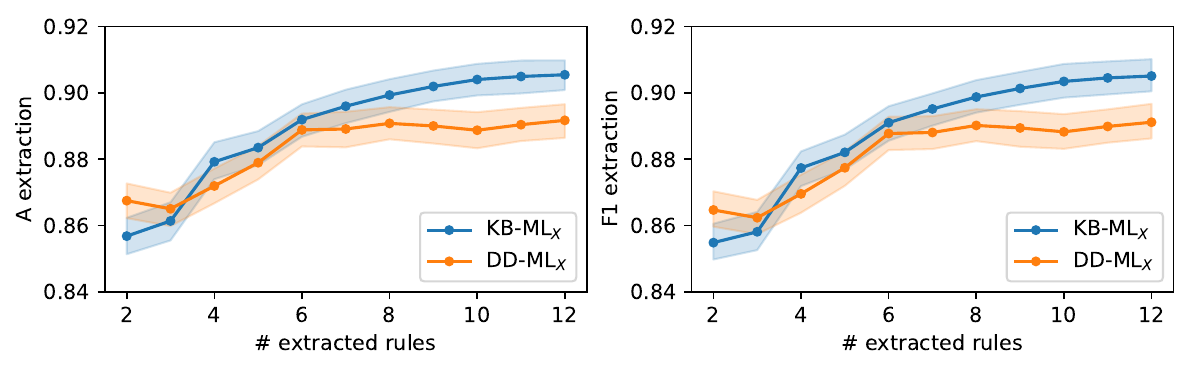}
        \caption{Accuracy and F1-score for extracted rule sets.}
    \end{subfigure}
    \begin{subfigure}{\textwidth}
        \centering
        \includegraphics[width=.9\linewidth]{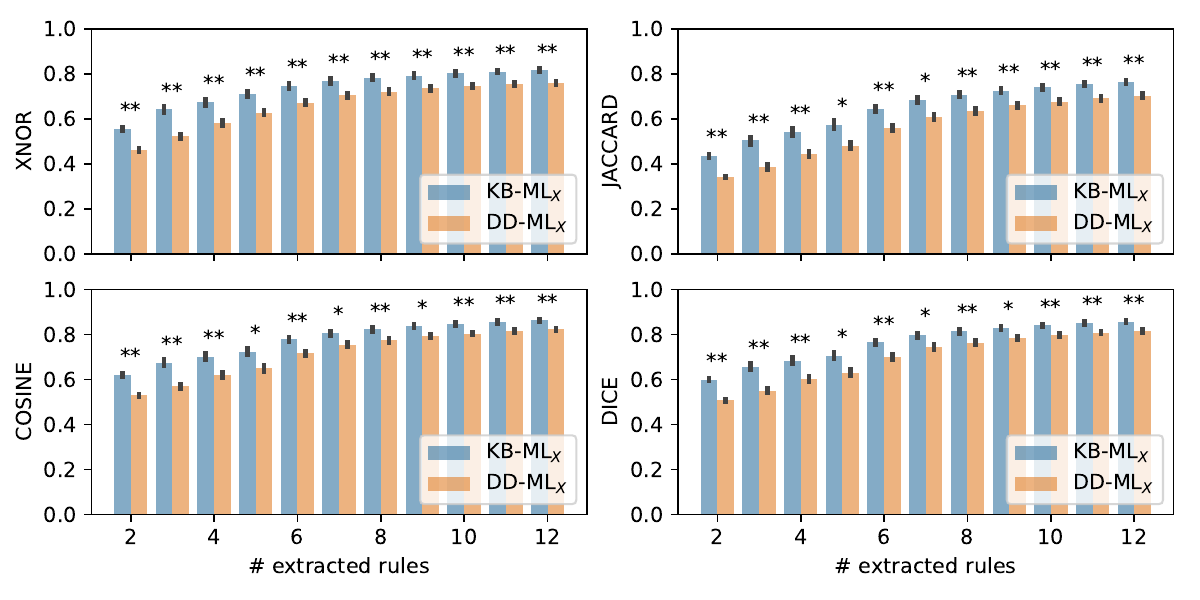}
        \caption{Explanation similarity metrics for explanation adherence.}
    \end{subfigure}    
    \begin{subfigure}{\textwidth}
        \centering
        \includegraphics[width=.9\linewidth]{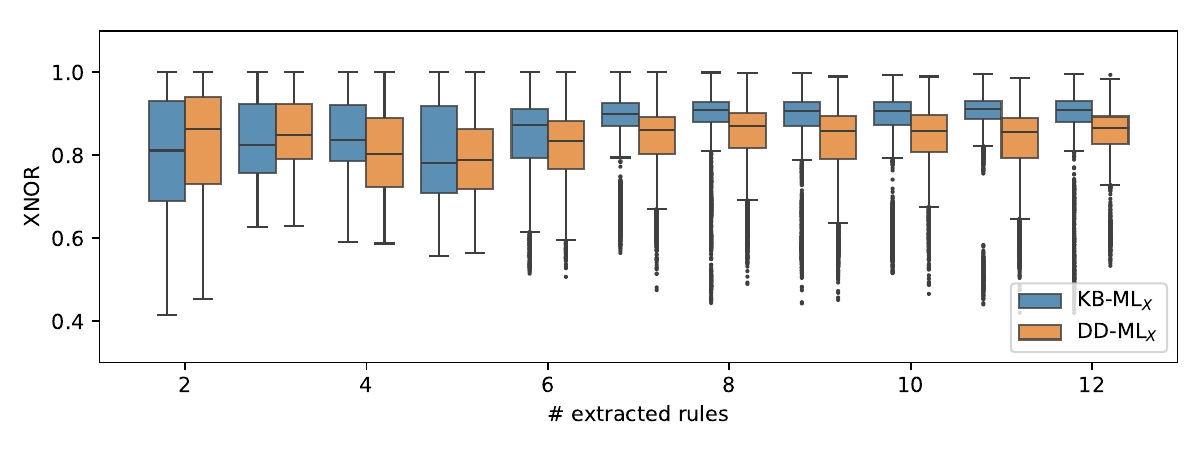}
        \caption{Explanation similarity metrics for model explanation robustness.}
    \end{subfigure}
 \caption{Average accuracy (A) and F1-score (F1) with 95\% confidence intervals of the rule sets DD-ML$_X$ and KB-ML$_X$, extracted from the data-driven models (DD-ML) or integrated models (KB-ML) using CART, with a varying number of rules extracted from 2 to 12.
 (b) Explanation similarity metrics (leveraging XNOR, Jaccard, Cosine, and Dice similarities) computed between the protocol and either DD-ML$_X$ or KB-ML$_X$ across 100 iterations, on all samples that can be predicted by all rule sets. (*) and (**) above the bar plots indicate significant differences between the values for the corresponding metric in DD-ML$_X$ and KB-ML$_X$ at a significance level of 0.05 and 0.01, respectively. 
 (c) Explanation similarity metrics for robustness evaluation, leveraging XNOR similarities, to evaluate similarities across the 100 instances of DD-ML$_X$ and similarly for KB-ML$_X$.
 \label{fig:cart_depth}}
 \end{center}
\end{figure}

The model incorporating domain knowledge also offers explanations that better align with the underlying reasoning of the knowledge base. Given the black-box nature of both the data-driven and integrated neural networks, explanations for each prediction are provided via surrogate rule sets, with a number of rules varying from 2 to 12, serving as approximations of the model's decision-making process. The surrogate models KB-ML$_X$ and DD-ML$_X$ closely mirror the behaviour of the black-box models, reporting accuracy and F1 scores consistently above 0.85 across all rule set sizes, as shown in Figure~\ref{fig:cart_depth}a. 

Explanation similarity metrics computed over samples with prediction for all considered models (KB, DD-ML, KB-ML) reveal that the similarity of KB-ML$_X$ to the knowledge base consistently exceeds that of DD-ML$_X$ across all similarity metrics and for every number of rules considered (Figure~\ref{fig:cart_depth}b). These differences are statistically significant across all metrics and rule set sizes. Notably, for the XNOR similarity, these differences maintain statistical significance at the 0.01 level across all rule set sizes, emerging as the most effective approach for capturing the impact of integration on improving explanation similarity to the established protocol. This is unsurprising, as the other similarities tend to give more emphasis to the overlapping of 1 values between the two local explanation vectors, while XNOR similarly accounts for the overlapping of 1s and 0s. This is desirable, as a 1 (meaning satisfied condition) in this context is as relevant as a 0 (i.e., unsatisfied condition).
%
%
Explanation similarity metrics computed over samples for which all considered models make a correct prediction verify that, with predictions being equal, explanations of the integrated model remain closer to the protocol than those of the data-driven model. In this analysis, a similar pattern is observed, with explanation similarity being greater for the integrated model across all metrics and numbers of rules, with differences statistically significant at the 0.01 level for XNOR and at the 0.05 level for all others.

Finally, the examination of explanation similarity across 100 instances of models trained via the 10-times 10-fold cross-validation, depicted in Figure~\ref{fig:cart_depth}c, reveals that similarity among KB-ML$_X$ rule sets is comparable to that of DD-ML$_X$ for rule sets comprising up to 5 rules. However, it surpasses that of DD-ML$_X$ for rule sets with more rules, which also have greater fidelity with the black-box model.
These findings demonstrate that the integrated model generates explanations that not only are more aligned with domain knowledge but are also more robust compared to the fully data-driven model for larger and more accurate rule sets, and that the proposed explanation similarity strategy is instrumental in evaluating this crucial aspect.

This approach presents notable advantages compared to strategies that rely solely on rules as global explanations for the model. Leveraging local explanations offers a more nuanced and fine-grained evaluation of model explanations, reflecting the structure of the data and providing more context-aware insights into the model's inner workings, which is particularly relevant in clinical settings.
The proposed approach offers several additional benefits. It can be applied to both numerical and categorical features. Instead of discretising data first and then building rule sets, it uses rule thresholds for data discretisation, eliminating the need of prior knowledge of relevant intervals. Furthermore, it provides a representation that automatically performs feature selection, excluding variables not present in the rules from the vector representation. It also accommodates variables included in other rule sets but not present in the knowledge base. In this scenario, rule sets with conditions on variables not accounted for by the knowledge base will have certain non-overlapping vector regions with the base and will likely record a lower score. Conversely, rule sets using the same features as the base will have greater opportunities for vector overlap and will typically yield higher scores. Lastly, it has a low computational cost, with similarity computation growing linearly with the number of samples, unlike methods that compute pairwise rule similarities, which grow quadratically with the number of rules.


\section{Conclusions and future work}
This study introduces novel metrics to evaluate the adherence of models to established protocols in terms of accuracy and explanation of predictions. Through comparative analysis on a benchmark dataset, we illustrate that models incorporating protocol knowledge exhibit superior alignment with established practices, making them more suitable for integration into clinical decision-making processes. 

In future research, we aim to extend this investigation to other datasets, retrieving the corresponding domain knowledge either by translating established protocols into rules or by consulting clinicians to encode that knowledge.
Having demonstrated adherence to the clinical protocol across different datasets and clinical applications, we also plan to consult respective experts to verify that the trained ML model is trustworthy also outside the domain of application of a protocol, by evaluating whether the learning criteria align with clinicians' judgement in borderline cases. Additionally, we plan to validate the proposed approach using other automatic rule extraction algorithms, including those based on fuzzy logic, such as neuro-fuzzy models. Finally, we intend to enhance the explanation similarity metrics by scaling intervals based on their length or the number of samples within them, rather than assigning binary values.

\paragraph{Availability of data and code}
The dataset analysed is publicly available (\url{https://www.kaggle.com/datasets/uciml/pima-indians-diabetes-database}), and the code to replicate the experiments can be found in the GitHub repository (\url{https://github.com/ChristelSirocchi/XAI-similarity}).

\bibliography{references}

\end{document}